# Anomaly Detection in Driving by Cluster Analysis Twice

Chung-Hao Lee, Yen-Fu Chen

**Abstract**— Events deviating from normal traffic patterns in driving, anomalies, such as aggressive driving or bumpy roads, may harm delivery efficiency for transportation and logistics (T&L) business. Thus, detecting anomalies in driving is critical for the T&L industry. So far numerous researches have used vehicle sensor data to identify anomalies. Most previous works captured anomalies by using deep learning or machine learning algorithms, which require prior training processes and huge computational costs. This study proposes a method namely Anomaly Detection in Driving by Cluster Analysis Twice (ADDCAT) which clusters the processed sensor data in different physical properties. An event is said to be an anomaly if it never fits with the major cluster, which is considered as the pattern of normality in driving. This method provides a way to detect anomalies in driving with no prior training processes and huge computational costs needed. This paper validated the performance of the method on an open dataset.

―――――――――― ◆ ――――――――――

## 1 INTRODUCTION

Vehicles are now essential in the Transportation and Logistics (T&L) industry. The market of the global T&L industry is estimated to reach US$4.6 trillion [1], [2]. Vehicle status in driving, the vehicle status that is influenced by every event occurring during driving, is critical for T&L corporations to monitor delivery efficiency and evaluate if a driver or a route is appropriate for the delivery task. Any event deviating from normal traffic patterns, as we called an anomaly in this study, may deteriorate delivery efficiency and vehicle status[3], [4]. Anomaly detection, therefore, is vital to ensure delivery efficiency and transportation activities. Anomalies comprise two major events, aggressive driving behaviors and bad road conditions [5]. Sensors are efficient tools to monitor and track aggressive driving behaviors and bad road conditions [6]–[8].

As more intelligent technology (5G, IoT and cloud) empower vehicles, the number of sensors on a vehicle has also increased [9]. G-sensor (gsen) and gyroscope (gyro) are two important sensors installed in a vehicle to measure physical properties during driving, such as: acceleration, orientation, and angular velocity. As G-sensors and gyroscopes are made more and more sophisticated, the quality of data they provide is improving. A great number of studies used data collected from G-sensor and gyroscope to find anomalies in driver behavior analyzing, road condition detecting and vehicle status estimating [10]–[12].

Machine learning (ML) now the most popular approach has been applied to sensor data analysis [13], [14]. The methods of anomaly detection in most studies can be categorized into 3 groups: threshold-based method, ML-based method, and deep learning (DL)-based method. Each group of methods has its pros and cons. For example, a threshold-based method does not require a training process but is susceptible to noises. A ML-based method generally takes less computational cost and may not need a training process (depending on which model uses) compared to DL algorithms, but some models need labeled data. A DL-based method can achieve relatively high accuracy, but it requires huge computational cost and training time and a large sample size. Each group of methods has many publications showing multitudinous algorithms to detect anomalies during driving [10]. Most studies used DL-based or ML-based methods to spot anomalies in driving. [15]–[19]. Even though these published algorithms have high accuracy, they require prior training processes and huge computational costs, which may not be suitable for some T&L companies.

Cluster analysis is the way to group a set of objects in the same concept that they are more similar to each rather than other groups. Compared to DL-based methods and some ML-based methods, clustering does not require a prior training process and huge computational costs. All these advantages make clustering an appropriate method to detect anomalies in real-world driving. For example, Martinelli et al. applied cluster analysis to identify driver aggressiveness [20]. Rajput et al. monitored road conditions by using unsupervised clustering methods (k-means and Self-Organizing Map) [21]. Ranjith et al. detected anomalies for traffic video surveillance by using Density-Based Spatial Clustering of Applications with Noise (DBSCAN) [22].

Multifarious clustering algorithms have been published. Each of them has its fit to specific applications [23]–[25]. In the case of anomaly detection in driving, we reckoned Hierarchical Density-Based Spatial Clustering of Applications with Noise (HDBSCAN) is well suited because of two reasons. First, unlike k-means, HDBSCAN doesn't need to predefine the number of clusters, which is

―――――――――――
- *Chung-Hao Lee is with the Wistron Corporation, New Taipei City, TW 22181. E-mail: chunghao_lee@wistron.com.*
- *Yen-Fu Chen is with the Wistron Corporation, New Taipei City, TW 22181. E-mail: andy_yf_chen@wistron.com.*

hard to know in advance of the driving. Second, due to several factors affecting interaction, the densities of sensor data during driving vary. HDBSCAN, compared to DBSCAN and OPTICS, has more capability to find clusters of varying densities [26], [27].

In this study, we developed a method namely Anomaly Detection in Driving by Cluster Analysis Twice (ADDCAT) that used HDBSCAN twice to detect traffic anomalies. We applied ADDCAT to the open datasets to validate its performance.

## 2 DATA

The open dataset we used was collected by Bhatt et al.. In this dataset, all of the data was collected on a 2007 Toyota Prius with about 100,000 miles. Sensor data (G-sensor and gyroscope) were collected by iPhone 6Ss. The sample rate is 5Hz [12].

The dataset has latitude, longitude, Unix timestamp, vehicle speed (m/s), gsen[X, Y, Z] (m/s$^2$), and gyro[X, Y, Z] (rad/s). The definition of the axes of a vehicle shows in Figure 1. The X-axis defines the direction of forwarding and backwarding. The Y-axis defines the direction of horizontal movements. The Z-axis defines the direction of vertical movements.

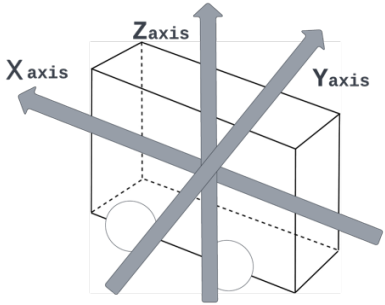

Figure 1. Axes of a vehicle

## 3 METHODS

All data analyses were conducted in Python (version 3.10.8) on JupyterLab (version 3.5.0) on MacBook Pro (13-inch 2019, 1.4GHz Quad-Core Intel Core i5, 16GB 2133MHz LPDDR3, Intel Iris Plus Graphics 645 1536MB, macOS Big Sur 11.5.2). The Python packages used but not limited to NumPy, pandas, matplotlib, seaborn, scikit-learn, HDBSCAN [28] and Uniform Manifold Approximation and Projection (UMAP) [29].

We presented our research methods in two sections: data processing and clustering, as reported in Figure 2.

### 3.1 Windowing

We followed Bhatt et al.'s procedure to group 10 data points (2 seconds) into intervals and calculated aggregated statistics for each interval from individual statistics. We created 20 aggregate statistics (AS) for each interval: mean of speed, standard deviation (std) of speed, maximum (max) of gsen[X, Y, Z], minimum (min) of gsen[X, Y, Z], std of gsen[X, Y, Z], max of gyro[X, Y, Z], min of gyro[X, Y, Z] and std of gyro[X, Y, Z]. We called each record

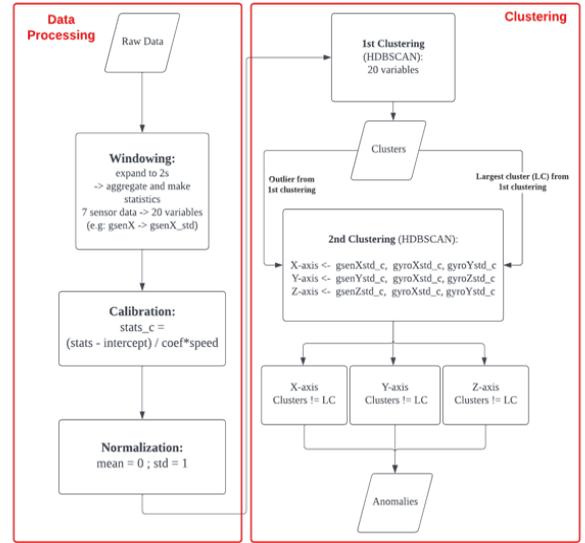

Figure 2. Workflow of Anomaly Detection in Driving by Cluster Analysis Twice

of aggregated statistics an event.

### 3.2 Calibration

Speed can affect G-sensor detection, the higher the speed the higher the acceleration values [30]. As an example, Figure 3a shows the speed-dependency of the various AS. This dependency provides no benefit for classification because data at a low speed may display differently than points at a high speed. However, since theoretically anomalies can occur in equal likelihood at any speed, it is clear that speed cannot by itself be used to classify anomalies. Thus, in order to remove linear dependence on speed and form a new speed independent data set as shown in Figure 3b, we modified the method listed in Perttunen et al. 's report [31]. For each AS, we first fit linear equations where y is AS, x is the mean of speed, a is a coefficient of a linear equation, and b is a y-intercept of a linear equation.
$$y = ax + b$$
Then, we formed a new linearly (speed) independent data set, where c is calibrated AS.
$$c = (y - b) / ax$$

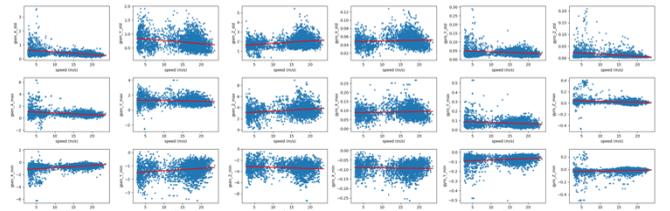

Figure 3a. Various aggregated statistics show speed-dependency before calibration

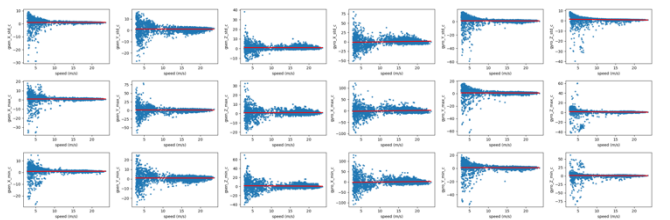

Figure 3b. Speed-dependency was removed from various aggregated statistics after calibration

### 3.3 Clustering

Before clustering, we normalized all 20 AS to the distribution in means equal to zero and std equal to one. We then applied HDBSCAN clustering algorithms twice. In the first clustering, we inputted all 20 normalized and calibrated AS to HDBSCAN algorithms, and we got clusters. To find anomalies in the X, Y, and Z axes, we did the second clustering. We applied HDBSCAN algorithms to each X, Y and Z axis of both separated groups, outliers, and the largest cluster from the first clustering result. In X-axis clustering, the inputted variables were normalized and calibrated std of gsenX, std of gyroX and std of gyroY. In Y-axis clustering, the inputted variables were normalized and calibrated std of gsenY, std of gyroX and std of gyroZ. In Z-axis clustering, the inputted variables were normalized and calibrated std of gsenZ, std of gyroX and std of gyroY.

After clustering twice, if an event was categorized to the largest clusters in all axes clustering, we defined it as a normal event, otherwise, an anomaly.

## 4 RESULTS

Figure 4 indicates the result of the first clustering on the open dataset. There are eight events in label 0 of the first clustering, twelve events in label 1, 483 events in label 2, and 374 events in label -1, which means outliers in this clustering. The largest cluster in the first clustering is label 2. Events were separated by speed after the first clustering. Cluster label 2 has a speed mean of 10.37m/s, while Cluster label -1 has a speed mean of 4.49m/s. Speed std shows no significant difference between cluster label 2 and label -1, where 0.61 and 0.74, respectively.

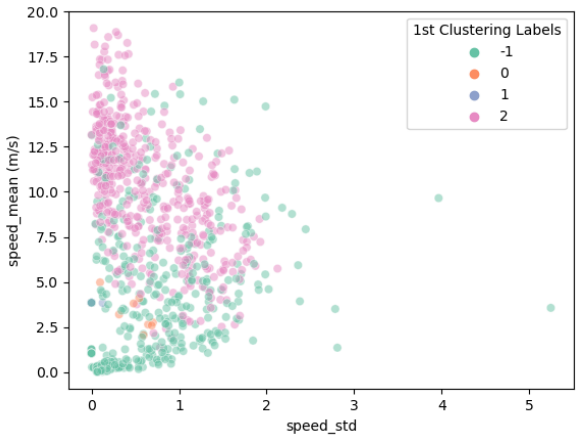

Figure 4. The result of the first clustering colored by cluster labels

Table 1 shows the result of the second clustering. Even though the second clustering generates various numbers of cluster labels in different axes, all axes can be separated into two major groups, the largest cluster, and outliers.

TABLE 1
NUMBER OF EVENTS IN THE RESULT OF THE SECOND CLUSTERING

| Cluster Labels of 2nd clustering | The Largest Cluster in 1st clustering | | | Outliers in 1st clustering | | |
|---|---|---|---|---|---|---|
| | X axis | Y axis | Z axis | X axis | Y axis | Z axis |
| Label -1 | 73 | 134 | 15 | 46 | 82 | 103 |
| Label 0 | 5 | 9 | 4 | 13 | 6 | 11 |
| Label 1 | 5 | 6 | 464 | 8 | 4 | 6 |
| Label 2 | 400 | 334 | – | 307 | 7 | 254 |
| Label 3 | – | – | – | – | 275 | – |

After clustering twice, ADDCAT detected 315 events as anomalies. Most anomalies were at low speeds. But some high and middle-speed events were also identified as anomalies by ADDCAT as shown in Figure 5a. Figure 5b reports most events recognized as normal were located in the central area of clusters, while anomalies were scattered across the peripheral area of clusters in all six slice plots. In Figure 5c, events recognized as normals expressed more concentrated patterns and events recognized as anomalies exhibited more linear looking from the perspective of calibrated AS compared to the perspective of uncalibrated AS shown in Figure 5b.

We used UMAP, a dimension reduction technique that can be used for visualization, to display the two-dimensional distribution of the open datasets showing in Figure 5d. Some potholes were located on the edge of the main group, while most were stood closer to the interior of the main group. ADDCAT identified more interior potholes and fewer peripheral ones. Events after UMAP treating were gathered into two groups, the larger main group and the smaller horn-like group located on the top left in Figure 5d (blue circle). All events at the smaller horn-like group were identified as anomalies. Confusion Matrix (Figure 6) indicates that ADDCAT achieved 0.62 accuracy on the open dataset and identified 31 out of 79 potholes as anomalies.

Table 2 reports comparison between true positives and false positives. True positives were events that were potholes and were identified as anomalies. False positives were events that were potholes but were not identified as anomalies. There were 31 true positives and 48 false positives. The average speed of true positives was 42.55kph, while the average speed of false positives was 37.78kph. Most of the statistics of true positives were greater than those of false positives.

## 5 DISCCUSIONS

In this report, we introduced ADDCAT, a method to detect anomalies in driving. The core concept of ADDCAT is similarity gathering. Similar driving events are supposed to have similar patterns. Usually, patterns of normality take the majority of all patterns in a driving route. Thus, if data accumulates enough, ADDCAT, ideally, can detect any types of anomalies in driving, such as traffic congestion, harsh braking, harsh cornering, speed bump, potholes, and other anomalies, because patterns of anomalies are different from patterns of normality. Compared to previous published algorithms for driving anomalies detection, ADDCAT does not require a training process, huge computational cost, nor the number of clusters before hitting the road.

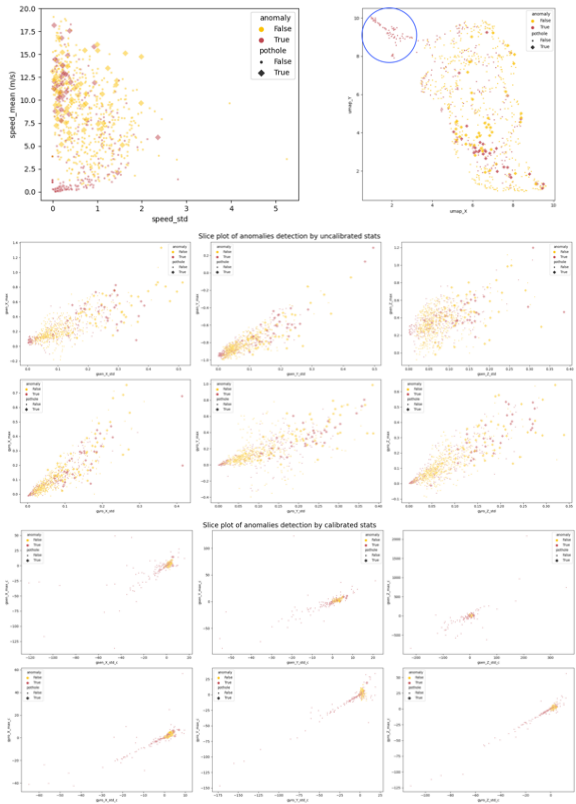

Figure 5a. (top-left) The result of the anomaly detection (colored by anomaly)
Figure 5b. (middle) Slice plot of anomaly detection by uncalibrated AS (colored by anomaly)
Figure 5c. (bottom) Slice plot of anomaly detection by calibrated AS (colored by anomaly)
Figure 5d. (top-right) UMAP of the anomalies detection by ADDCAT (colored by anomaly, shaped by pothole). The horn-like group in the blue circle is separated from the larger main group.

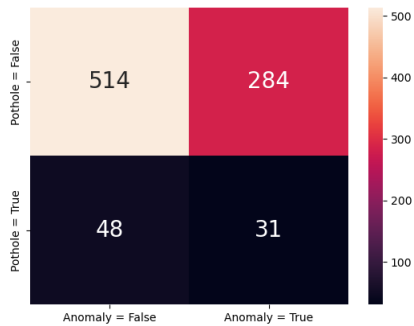

Figure 6. Confusion Matrix of the anomaly detection by ADDCAT

TABLE 2
COMPARISON BETWEEN TRUE POSITIVE AND FALSE POSITIVE

| Statistics | True Positive | False Positive |
|---|---|---|
| speed (kph) | 42.55 | 37.78 |
| gsenZstd_c | 2.64 | 2.39 |
| gsenZmax_c | 4.65 | 4.17 |
| gsenZmin_c | 1.93 | 1.95 |
| gyroXstd_c | 2.04 | 1.98 |
| gyroXstd_c | 2.10 | 2.13 |

From the result of the first clustering, we can deduce that speed is the most important factor among all factors to separate clusters. The process to remove speed-dependency, calibration, therefore, is the key to make other AS more impactful on cluster dissociation.

Figure 5a indicates ADDCAT could identify potholes from low speed to high speed, but it did not perform well to those potholes with larger speed std. Even though ADDCAT was able to detect potholes with larger AS, from Figure 5b, we can see that ADDCAT failed to recognize the most extreme events in slice plots of gsenX, gyroY and gyroZ. ADDCAT did not catch those potholes because the AS of those potholes shrank toward zero after calibration. That means the AS values of unrecognized potholes after calibration were similar with the majority events. That's why ADDCAT failed to isolate unrecognized potholes from the major cluster showing in Figure 5c.

The speed mean of the horn-like group in Figure 5d was 2.8 kph and the std of speed was 0.19. The vehicle with this low speed indicates that it was probably ready to stop for traffic lights or it was stuck in traffic congestion. Either way belonged to the pattern of normality. But because the patterns of stop for traffic lights or traffic congestion are different from regular driving (speed and sensor data), ADDCAT picked these patterns out and set them apart from the main group. If we consider these patterns as patterns of normality, the number of false negatives can reduce to 193 and the accuracy can rise to 0.73.

Because we have removed speed-redundancy, although the statistics of false positive potholes were large before calibration, the statistics of these potholes were calibrated to the numbers that were similar with the statistics of the normalities. This may imply the large statistics numbers generated by sensors only occur when a car hits these potholes at a fast speed. Thus, the size and the depth of these potholes may not be huge.

ADDCAT uses similarity gathering to separate driving events. There are only 877 events in the open datasets, which may not have enough data for ADDCAT to form a cluster for some rare driving events or anomalies. If abundant data are provided to ADDCAT, more similar patterns of driving events can form clusters. Thus, the detection capability of ADDCAT can be enhanced.

ADDCAT does not need prior training nor deciding number of clusters. This method not only can help the T&L industry evaluate vehicle status, but also can help tech companies develop their autonomous driving technology. Most autonomous driving technology are built on DL algorithms, which require a cornucopia of labeled data to train. It is costly, time-consuming and manual labor intensive to obtain labeled data[32], [33]. ADDCAT provides a new approach that can reduce the cost of labeling data, and then help tech companies in saving costs and shortening time to develop an autonomous driving technique.

## 6 CONCLUSIONS

In this study, we presented a method namely Anomaly Detection in Driving by Cluster Analysis Twice (ADDCAT) and validated it in an open dataset. We accomplished the data processing to reflect the real physical effect caused by potholes by removing speed-dependency. We use HDBSCAN for the clustering process. We applied the first cluster analysis to processed sensor data to separate them based on their speeds. We applied the second cluster analysis to the clusters from the result of first clustering to identify anomalies.

We successfully provided demand sides with an anomaly detection method in low cost, less computational capacity needed, no prior training process requirement and no number of cluster decisions beforehand. In the future, ADDCAT not only can help the T&L industry evaluate traffic conditions and vehicle status, but also can help tech companies to label traffic events to expedite development of autonomous driving.

## FUNDING

This research received no external funding.

## ACKNOWLEDGMENT

The authors wish to thank Yung-Hui Hsu and Peng-Cheng Chen for supporting.

## REFERENCES

[1] A. Tipping and P. Kauschke, "Shifting patterns: The future of the logistics industry.," *Price Waterhouse Coopers, Phoenix*, 2016.

[2] V. P. Gao, H.-W. Kaas, D. Mohr, D. Wee, and T. Möller, "Automotive revolution – perspective towards 2030," McKinsey & Company, Jan. 2016.

[3] B. Bose, J. Dutta, S. Ghosh, P. Pramanick, and S. Roy, "D&rsense: detection of driving patterns and road anomalies," in *2018 3rd International Conference On Internet of Things: Smart Innovation and Usages (IoT-SIU)*, Feb. 2018, pp. 1–7, doi: 10.1109/IoT-SIU.2018.8519861.

[4] O. Köpüklü, J. Zheng, H. Xu, and G. Rigoll, "Driver Anomaly Detection: A Dataset and Contrastive Learning Approach," *arXiv*, 2020, doi: 10.48550/arxiv.2009.14660.

[5] Y. Al-Wreikat, C. Serrano, and J. R. Sodré, "Driving behaviour and trip condition effects on the energy consumption of an electric vehicle under real-world driving," *Appl. Energy*, vol. 297, p. 117096, Sep. 2021, doi: 10.1016/j.apenergy.2021.117096.

[6] N. AbuAli, "Advanced vehicular sensing of road artifacts and driver behavior," in *2015 IEEE Symposium on Computers and Communication (ISCC)*, Jul. 2015, pp. 45–49, doi: 10.1109/ISCC.2015.7405452.

[7] Y. Li, F. Xue, L. Feng, and Z. Qu, "A driving behavior detection system based on a smartphone's built-in sensor," *Int. J. Commun. Syst.*, vol. 30, no. 8, p. e3178, May 2017, doi: 10.1002/dac.3178.

[8] N. Kalra, G. Chugh, and D. Bansal, "Analyzing driving and road events via smartphone," *IJCA*, vol. 98, no. 12, pp. 5–9, Jul. 2014, doi: 10.5120/17233-7561.

[9] S. Abdelhamid, H. S. Hassanein, and G. Takahara, "Vehicle as a mobile sensor," *Procedia Computer Science*, vol. 34, pp. 286–295, 2014, doi: 10.1016/j.procs.2014.07.025.

[10] E. A. Martinez-Ríos, M. R. Bustamante-Bello, and L. A. Arce-Sáenz, "A Review of Road Surface Anomaly Detection and Classification Systems Based on Vibration-Based Techniques," *Appl. Sci.*, vol. 12, no. 19, p. 9413, Sep. 2022, doi: 10.3390/app12199413.

[11] K. B. Singh, M. A. Arat, and S. Taheri, "Literature review and fundamental approaches for vehicle and tire state estimation," *Vehicle System Dynamics*, pp. 1–23, Nov. 2018, doi: 10.1080/00423114.2018.1544373.

[12] U. Bhatt, S. Mani, E. Xi, and J. Z. Kolter, "Intelligent Pothole Detection and Road Condition Assessment," *arXiv*, 2017, doi: 10.48550/arxiv.1710.02595.

[13] F. Zantalis, G. Koulouras, S. Karabetsos, and D. Kandris, "A review of machine learning and iot in smart transportation," *Future Internet*, vol. 11, no. 4, p. 94, Apr. 2019, doi: 10.3390/fi11040094.

[14] Z. Elamrani Abou Elassad, H. Mousannif, H. Al Moatassime, and A. Karkouch, "The application of machine learning techniques for driving behavior analysis: A conceptual framework and a systematic literature review," *Eng. Appl. Artif. Intell.*, vol. 87, p. 103312, Jan. 2020, doi: 10.1016/j.engappai.2019.103312.

[15] J. M. Celaya-Padilla *et al.*, "Speed bump detection using accelerometric features: A genetic algorithm approach.," *Sensors*, vol. 18, no. 2, Feb. 2018, doi: 10.3390/s18020443.

[16] D. Luo, J. Lu, and G. Guo, "Road anomaly detection through deep learning approaches," *IEEE Access*, vol. 8, pp. 117390–117404, 2020, doi: 10.1109/ACCESS.2020.3004590.

[17] D. Dong and Z. Li, "Smartphone sensing of road surface condition and defect detection.," *Sensors*, vol. 21, no. 16, Aug. 2021, doi: 10.3390/s21165433.

[18] N. Silva, J. Soares, V. Shah, M. Y. Santos, and H. Rodrigues, "Anomaly Detection in Roads with a Data Mining Approach," *Procedia Computer Science*, vol. 121, pp. 415–422, 2017, doi: 10.1016/j.procs.2017.11.056.

[19] "Automatic Road Anomaly Detection Using Smart Mobile Device | NTU Scholars." https://scholars.lib.ntu.edu.tw/handle/123456789/358923 (accessed Dec. 08, 2022).

[20] F. Martinelli, F. Mercaldo, V. Nardone, A. Orlando, and A. Santone, "Cluster analysis for driver aggressiveness identification," in *Proceedings of the 4th International Conference on Information Systems Security and Privacy*, Jan. 2018, pp. 562–569, doi: 10.5220/0006755205620569.

[21] P. Rajput, M. Chaturvedi, and V. Patel, "Road condition monitoring using unsupervised learning based bus trajectory processing," *Multimodal Transportation*, vol. 1, no. 4, p. 100041, Dec. 2022, doi: 10.1016/j.multra.2022.100041.

[22] R. Ranjith, J. J. Athanesious, and V. Vaidehi, "Anomaly detection using DBSCAN clustering technique for traffic video surveillance," in *2015 Seventh International Conference on Advanced Computing (ICoAC)*, Dec. 2015, pp. 1–6, doi: 10.1109/ICoAC.2015.7562795.

[23] A. Ali *et al.*, "Systematic review: A state of art ML based clustering algorithms for data mining," in *2020 IEEE 23rd In-*

*ternational Multitopic Conference (INMIC)*, Nov. 2020, pp. 1–6, doi: 10.1109/INMIC50486.2020.9318060.

[24] R. Xu and D. Wunsch, "Survey of clustering algorithms.," *IEEE Trans. Neural Netw.*, vol. 16, no. 3, pp. 645–678, May 2005, doi: 10.1109/TNN.2005.845141.

[25] M. Verma, M. Srivastava, N. Chack, A. K. Diswar, and N. Gupta, "A Comparative Study of Various Clustering Algorithms in Data Mining," *IJERA*, vol. 2, no. 3, pp. 1379–1384, May 2012.

[26] R. J. G. B. Campello, D. Moulavi, and J. Sander, "Density-Based Clustering Based on Hierarchical Density Estimates," in *Advances in Knowledge Discovery and Data Mining*, vol. 7819, J. Pei, V. S. Tseng, L. Cao, H. Motoda, and G. Xu, Eds. Berlin, Heidelberg: Springer Berlin Heidelberg, 2013, pp. 160–172.

[27] L. McInnes and J. Healy, "Accelerated hierarchical density based clustering," in *2017 IEEE International Conference on Data Mining Workshops (ICDMW)*, Nov. 2017, pp. 33–42, doi: 10.1109/ICDMW.2017.12.

[28] L. McInnes, J. Healy, and S. Astels, "hdbscan: Hierarchical density based clustering," *JOSS*, vol. 2, no. 11, p. 205, Mar. 2017, doi: 10.21105/joss.00205.

[29] T. Sainburg, L. McInnes, and T. Q. Gentner, "Parametric UMAP embeddings for representation and semi-supervised learning," *arXiv*, 2020, doi: 10.48550/arxiv.2009.12981.

[30] T. Haniszewski and A. Michta, "Preliminary studies of vertical acceleration of a passenger car passing through the speed bump for various driving speeds," *Transport Problems*, vol. 14, no. 1, pp. 23–34, 2019, doi: 10.21307/tp.2019.14.1.3.

[31] M. Perttunen *et al.*, "Distributed road surface condition monitoring using mobile phones," in *Ubiquitous intelligence and computing*, vol. 6905, C.-H. Hsu, L. T. Yang, J. Ma, and C. Zhu, Eds. Berlin, Heidelberg: Springer Berlin Heidelberg, 2011, pp. 64–78.

[32] J. Kim, J. Ju, R. Feldt, and S. Yoo, "Reducing DNN labelling cost using surprise adequacy: an industrial case study for autonomous driving," in *Proceedings of the 28th ACM Joint Meeting on European Software Engineering Conference and Symposium on the Foundations of Software Engineering*, New York, NY, USA, Nov. 2020, pp. 1466–1476, doi: 10.1145/3368089.3417065.

[33] A. Behl, K. Chitta, A. Prakash, E. Ohn-Bar, and A. Geiger, "Label efficient visual abstractions for autonomous driving," in *2020 IEEE/RSJ International Conference on Intelligent Robots and Systems (IROS)*, Oct. 2020, pp. 2338–2345, doi: 10.1109/IROS45743.2020.9340641.

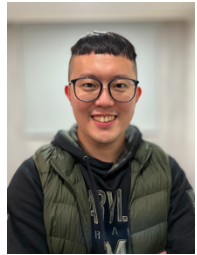

**Chung-Hao Lee** received the bachelor's degree from National Tsing Hua University, Taiwan and master's dual degrees in MBA and Master of Science in Information Systems from University of Maryland, College Park. He received an outstanding graduate student award nomination at UMD in 2022. Chung-Hao is currently a data scientist at Wistron. His main research interests include machine learning application on automobile, manufacturing, biotech, and sport analysis.
https://www.linkedin.com/in/lch99310/

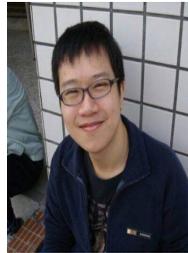

**Yen-Fu Chen** a staff software engineer at Wistron. a decade of the experiment for designing computer system architecture and system problem-solving, recently He focused on landing AIOps MLOps inside the corporation, At 2021 He led Wistron's first cloud-native AI inference platform for Taipei's MRT and Taiwan railway system, 2022 He led an AI start-up department in Wistron.
https://www.linkedin.com/in/%E9%99%B3-%E5%BD%A5%E7%94%AB-6a7b8716/